\documentclass[nohyperref]{article}

\usepackage{microtype}
\usepackage{booktabs} 

\usepackage[utf8]{inputenc}
\usepackage[round]{natbib}

\usepackage{hyperref}
\usepackage{listings}
\usepackage[inline, shortlabels]{enumitem}
\usepackage{url}
\usepackage{soul, color}
\usepackage{lipsum}
\usepackage{xfrac}

\usepackage{amsmath, amssymb, amsthm}
\usepackage{mathtools}
\usepackage[ruled, vlined]{algorithm2e}
\usepackage{algpseudocode}

\usepackage{graphicx, subfigure}
\usepackage{multirow, longtable, diagbox, setspace, rotating}
\usepackage{wrapfig, float, caption}
\setlist{topsep=0pt, leftmargin=*}
 \usepackage[accepted]{icml2022}

\usepackage[capitalize,noabbrev]{cleveref}

\theoremstyle{plain}

\theoremstyle{definition}

\theoremstyle{remark}

\usepackage[textsize=tiny]{todonotes}

\icmltitlerunning{ModLaNets: Learning Generalisable Dynamics via Modularity and Physical Inductive Bias}

\begin{document}

\twocolumn[
\icmltitle{ModLaNets: Learning Generalisable Dynamics via Modularity and Physical Inductive Bias}



\icmlsetsymbol{equal}{*}

\begin{icmlauthorlist}
\icmlauthor{Yupu Lu}{uni,comp}
\icmlauthor{Shijie Lin}{uni,comp}
\icmlauthor{Guanqi Chen}{uni}
\icmlauthor{Jia Pan}{uni,comp}
\end{icmlauthorlist}

\icmlaffiliation{uni}{Department of Computer Science, The University of Hong Kong, Hong Kong SAR, China}
\icmlaffiliation{comp}{Centre for Garment Production Limited (TransGP), Hong Kong SAR, China}

\icmlcorrespondingauthor{Yupu Lu}{luyp16@connect.hku.hk}
\icmlcorrespondingauthor{Jia Pan}{jpan@cs.hku.hk}


\icmlkeywords{Generalisable Dynamics Learning, Modularity, Network Reuse, Physical Inductive Bias, Lagrangian Mechanics, Representation Learning}

\vskip 0.3in
]



\printAffiliationsAndNotice{}  

\begin{abstract}
Deep learning models are able to approximate one specific dynamical system but struggle at learning generalisable dynamics, where dynamical systems obey the same laws of physics but contain different numbers of elements (e.g., double- and triple-pendulum systems).
To relieve this issue, we proposed the Modular Lagrangian Network (ModLaNet), a structural neural network framework with modularity and physical inductive bias. 
This framework models the energy of each element using modularity and then construct the target dynamical system via Lagrangian mechanics. Modularity is beneficial for reusing trained networks and reducing the scale of networks and datasets. As a result, our framework can learn from the dynamics of simpler systems and extend to more complex ones, which is not feasible using other relevant physics-informed neural networks. 
We examine our framework for modelling double-pendulum or three-body systems with small training datasets, where our models achieve the best data efficiency and accuracy performance compared with counterparts. We also reorganise our models as extensions to model multi-pendulum and multi-body systems, demonstrating the intriguing reusable feature of our framework.
\end{abstract}

\section{Introduction}
\label{sec:intro}

Deep learning has been widely implemented in various kinds of physical problems such as flow control using reinforcement learning \citep{verma2018efficient}, nuclear learning and simulation \citep{pfau2020ab, hermann2020deep}, and super-resolution of combustion process \citep{zhao2020time}.
Even though deep learning models have demonstrated outstanding performance in various high-dimensional tasks, these models are still statistical models driven by data, so nonphysical behaviours are inevitably likely to occur in simulation tasks.

The fusion of physical knowledge and neural networks casts a light on this problem, which leads to physics-informed neural networks such as Hamiltonian Neural Networks (HNNs, \citealp{Greydanus2019deep}) or Lagrangian Neural Networks (LNNs, \citealp{cranmer2020lagrangian}). These models treated systems as a whole and regressed central quantities like Lagrangian or Hamiltonian to represent dynamics, which preserves part of physical properties such as energy conservation.

Despite the strengths, though deep learning models can approximate one whole system, they cannot generalise other systems with the same laws of physics. There are two weaknesses. First, users need to train different networks for these systems. For instance, considering the motions of multi-pendulum systems, the same laws govern their motions; the only difference is the number of pendulums. However, they cannot be generalised by one network. This example means these models only focus on approximating the physical quantities of one system. 
Second, as the scale of the system grows, quantities\textquotesingle \ forms become more complex, which requires larger networks and more data for training. The training difficulties will also increase. Table \ref{table:settings} provides an example of how the scale of these networks grows from single-pendulum to double-pendulum systems, where the size of networks grows by ten times.

\begin{table}[htbp]
\setstretch{0.8}
\vskip -0.2in
\caption{The scale of networks and datasets used for modelling pendulum systems\protect\footnotemark[3]. }
\vskip 0.05in
\label{table:settings}
\centering
\resizebox{.48\textwidth}{!}{
\begin{sc}
\begin{tabular}{lcc}
\toprule
 & Single Pendulum & Double Pendulum\\
\midrule
Parameters & 41.2k & 364.2k \\
\hline
Dataset Size & 2.25k & 307.2m \\
\bottomrule
\end{tabular}
\end{sc}
}
\vskip -0.1in
\end{table}
\setstretch{1}

\footnotetext[3]{Two cases are separately estimated from previous works \href{https://github.com/greydanus/hamiltonian-nn/tree/master/experiment-pend}{Hamiltonian Neural Networks} and \href{https://github.com/MilesCranmer/lagrangian_nns/blob/master/notebooks/DoublePendulum.ipynb}{Lagrangian Neural Networks}.}

To solve these problems, we introduce modularity into our framework. 
We use multiple light neural networks to model the properties, potential and kinetic energies of elements in the system. These modular networks represent the basic features of each element. Then we can use these elements to construct a group of systems by adding or removing elements. Furthermore, these trained networks can be shared with other elements for the same physical properties, which will reduce the scale of the network and dataset.

\textbf{Contributions.} First, we demonstrate why we should model elements in the systems rather than directly model the whole system from the view of Lagrangian mechanics. Second, we propose a novel modular and physical inductive bias framework to learn dynamical systems. We discuss how the framework models each element in a system and supports network reuse to construct different systems. Third, we show that our framework outperforms our counterparts with higher accuracy and better data and parameter efficiency in experiments. We also reveal the reuse feature as an extension to model complex systems. This feature opens up more possibilities for the role of deep learning models in simulation tasks.

\section{Related Works}
\label{sec:works}
\textbf{Neural Networks for Next-State Prediction.} Researchers have already built different neural networks for physics problems. For example, they implemented networks for chains motions \citep{de2018end} and multi-particles interactions \citep{battaglia2016interaction, kipf2018neural}, where researchers designed recursive neural networks or graphical neural networks to induce the relations and interactions among objects. These models directly predicted the system\textquotesingle s states, but they did not encode any physical knowledge and could achieve nonphysical results.

\textbf{Neural Networks for Dynamical System Reconstruction.} Researchers provided another way to train neural networks in the view of dynamical systems. They utilised the Pontryagin\textquotesingle s maximum principle to formulate continuous systems with optimality \citep{weinan2017proposal, li2017maximum, chen2018neural, chen2020learning, yu2021onsagernet}, and implemented neural networks to model system\textquotesingle s dynamics, which are ordinary differential equations (ODEs). The evolution of a system can be derived by integrating its ODE. This model can be easily implemented, but it becomes harder to regress complex dynamics involving derivatives and algebraic fractions.

\textbf{Neural Networks with Physical Inductive Biases.} To reduce the difficulty in training, researchers started to design physics-informed neural networks such as DeLaN, HNN, and LNN \cite{lutter2018deep, Greydanus2019deep, cranmer2020lagrangian}. These models, respectively, focused on reconstructing the Lagrangian and Hamiltonian to describe the physical systems. The target physical variables like accelerations could then be derived with physical prior. 

Furthermore, these works have been expanded. For example, networks were adjusted to model systems with image data \cite{toth2019hamiltonian, zhong2020unsupervised, allen2020lagnetvip} or with contact and constraints \cite{finzi2020simplifying, zhong2021extending}. For another application in control, since Lagrangian and Hamiltonian involve energy of the system, the representing networks were utilised as a component with the defined control policies. By doing this, researchers recovered dynamical systems, which corresponded to successful control policies \cite{zhong2019symplectic, lutter2019deep, roehrl2020modeling, zhong2020unsupervised}. Also, several works modelled Hamiltonian systems in the view of the symplectic map, with the utilisation of Bayesian optimisation, symplectic networks, or Poisson networks \cite{jin2020learning, jin2020sympnets, galioto2020bayesian, chen2021data}. \citealp{willard2020integrating} and \citealp{zhong2021benchmarking} provided related surveys on physics-based and energy-conserving neural networks for thorough overviews. 

Previous works aim at regressing systems as a whole. However, larger networks are needed for more complex systems, which means more data and challenges in training. 
Meanwhile, due to fixed input size, they struggle to extend to generalised dynamics, where dynamical systems contain different numbers or kinds of elements (e.g., double- and triple-pendulum systems).
To overcome these issues, we proposed our framework that supports separately establishing the physical model of each element, thereby reducing the scale of networks and datasets. Trained models can also be utilised in different systems as extensions.

\begin{figure*}[htbp]
\vskip -0.05in
\centering
\includegraphics[width=0.98\linewidth]{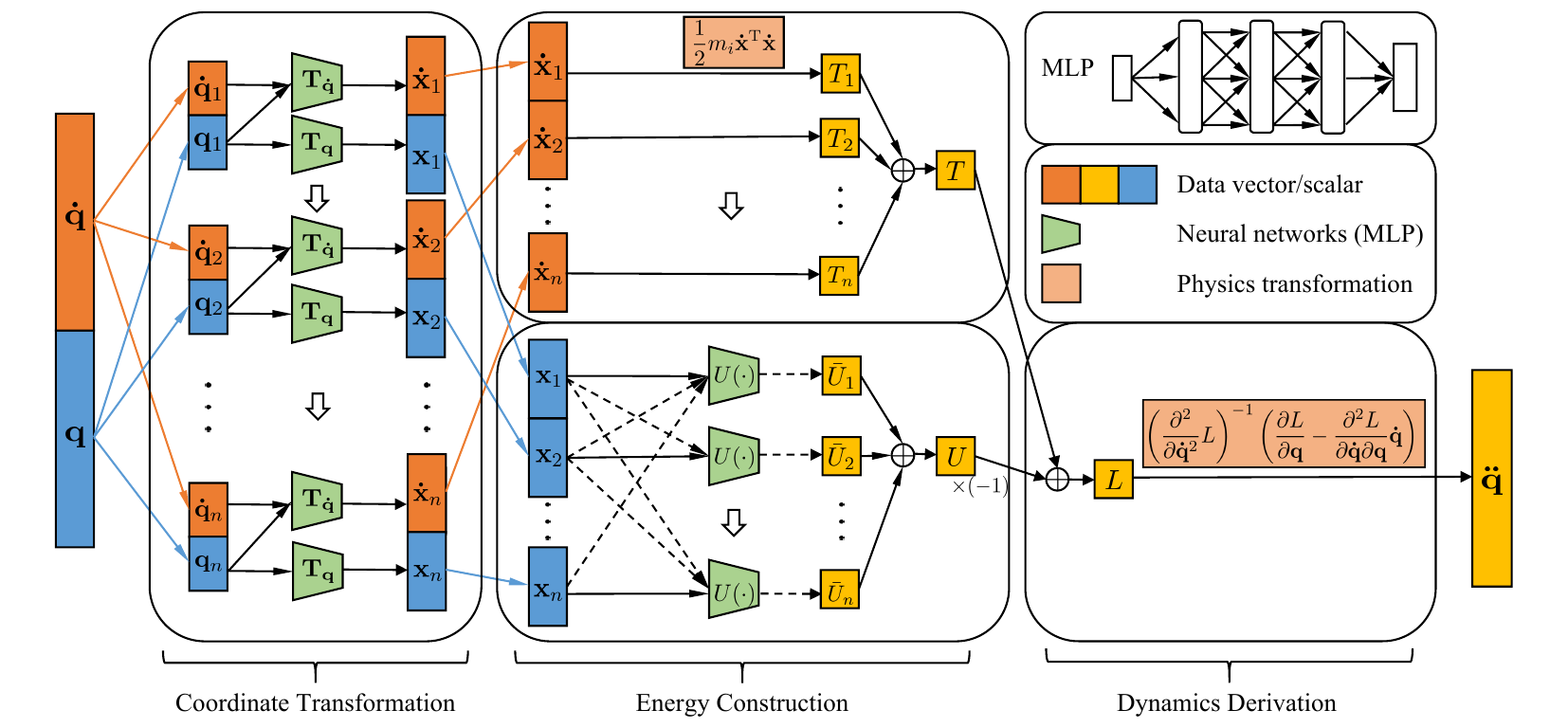}
\vskip -0.1in
\caption{The framework of ModLaNet. For each element, input ($\mathbf{\dot{q}}, \mathbf{{q}}$) will first be transformed to ($\mathbf{\dot{x}}, \mathbf{{x}}$) in the global coordinate system. After that, these global coordinates will be fed into networks to get each element\textquotesingle s potential and kinetic energy $(T, U)$. Here these networks are multilayer perceptrons (MLPs). The same type of elements shares their networks. The dashed arrows represent the changeable number of inputs to model different potential energies. At last, $T$ and $U$ will construct Lagrangian $L$ to derive the quantity $\mathbf{\ddot{q}}$.}
\label{fig:structure}
\vskip -0.1in
\end{figure*}

\section{Preliminaries}
\label{sec:pre}
The research target is to learn the evolution function of a dynamical system $f$ related to the system state ($\mathbf{q}, \mathbf{\dot{q}}$):
\begin{equation}
	\mathbf{\ddot{q}} = f_\theta(\mathbf{q}, \mathbf{\dot{q}}),
	\label{equ:system}
\end{equation}
where $\theta$ is a set of parameters, $\mathbf{q}, \dot{\mathbf{q}}$ and $\mathbf{\ddot{q}}$ represent the generalised position, velocity and acceleration, which are abstract concepts to represent systems\textquotesingle\ states like {[angle, angular velocity, angular acceleration]} based on coordinate systems.

In Lagrangian mechanics, function $f$ is determined by the Lagrangian $L$ of the system, which combines the kinetic energy $T$ and the potential energy $U$:
 \begin{equation}
	L(t, \mathbf{q}, \mathbf{\dot{q}}) \coloneqq T(t, \mathbf{q}, \mathbf{\dot{q}}) - U(t, \mathbf{q}, \mathbf{\dot{q}})\text{,}
	\label{equ:lag}
\end{equation}
In calculus of variations, the first-order of necessary conditions for weak extrema lead to \textit{the Euler-Lagrange equation} \citep{liberzon2011calculus}. When only $\mathbf{q}$ and $\mathbf{\dot{q}}$ are related to time $t$, we can expand Euler-Lagrange equation to:
\begin{equation}
\begin{split}
	\frac{\partial L(t, \mathbf{q}, \mathbf{\dot{q}})}{\partial \mathbf{q}} 
	&= \frac{\textrm{d}}{\textrm{d}t} \cfrac{\partial L(t, \mathbf{q}, \mathbf{\dot{q}})}{\partial \mathbf{\dot{q}}} \\
	&= \frac{\partial^2 {L(t, \mathbf{q}, \mathbf{\dot{q}})} }{\partial \mathbf{q}\partial {\mathbf{\dot{q}}}}  \frac{\textrm{d}\mathbf{q}}{\textrm{d}t} + \frac{\partial^2 {L}(t, \mathbf{q}, \mathbf{\dot{q}})}{\partial \mathbf{\dot{q}}^2} \frac{\textrm{d}\mathbf{\dot{q}}}{\textrm{d}t}
	\text{.}
	\label{equ:e-l}
\end{split}
\end{equation} 

From Equation (\ref{equ:e-l}), we then describe the evolution function of the dynamical system concerning $\mathbf{q}$ and $\mathbf{\dot{q}}$:
\begin{equation}
\begin{split}
\mathbf{\ddot{q}} = \left(\frac{\partial^2 {L}}{\partial \mathbf{\dot{q}}^2}\right)^{-1} \left(\frac{\partial {L}}{\partial \mathbf{q}} - \frac{\partial^2 {L} }{\partial \mathbf{q}\partial {\mathbf{\dot{q}}}} \mathbf{\dot{q}}\right) \text{,}
\label{equ:results}
\end{split}
\end{equation}
where each Lagrangian corresponds to a unique evolution function of a system such as a double-pendulum system. If this Lagrangian is directly recovered from data by a neural network \cite{cranmer2020lagrangian}, this network cannot represent general cases such as multi-pendulum systems. 
 
However, we can utilise the linear combination property of energy to construct Lagrangian in Equation (\ref{equ:lag}). A system\textquotesingle s kinetic energy is the combination of each element\textquotesingle s kinetic energy $T_{i}$: $T = \sum_{i} T_{i}$. And a system\textquotesingle s potential energy contains the potential energy between the element with the environment $U_i$ and the potential energy among elements $U_{ij}$. By evenly distributing $U_{ij}$ to elements $i, j$, we can achieve the combination related to each element: $U = \sum_i \bar{U}_i = \sum_i ({U}_i + \frac{1}{2}\sum_j U_{ij})$.

This hint suggests we should focus on elements inside a system rather than a whole system. After modelling energies of each element with modularity, we can construct different Lagrangians, thereby describing a group of systems.

\section{Method}
\label{sec:method}
In this section, we propose our framework called Modular Lagrangian Networks (ModLaNets), shown in Figure \ref{fig:structure}. This framework includes coordinate transformation, energy construction, and dynamics derivation. Using modularity, we reveal how each part works to model a system and reveal the feature of network reuse. Examples are also discussed to show how to set up the framework.

\textbf{Coordinate Transformation.} 
Transformation describes a system\textquotesingle s states from local coordinate systems to the global coordinate system, which is advantageous to unifying inputs and simplifying the construction of energy models. 

Local coordinate systems define how objects relate to others, describing how inputs are measured. For simplicity of symbols, we use $(\mathbf{q}, \mathbf{\dot{q}})$ to represent inputs in local coordinate systems. When these coordinate systems are constructed, we assume we know the relations between origins and elements. 

Global coordinate systems describe all objects in the space in a consistent form. We use $(\mathbf{x}, \mathbf{\dot{x}})$ to refer to positions and velocities in the global Cartesian system in our work. 

\begin{figure}[tbp]
\centering
\includegraphics[width=1\columnwidth]{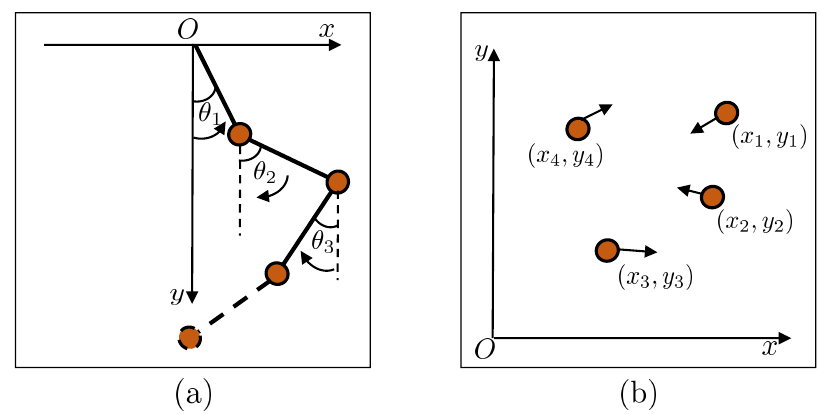}
\vskip -0.1in
\caption{Two dynamical systems, (a) multi-pendulum systems and (b) multi-body systems, for illustrations.}
\label{fig:Examples}

\end{figure}

One example is shown in Fig.\ref{fig:Examples}(a). For $i$-th pendulum, there exists a local polar coordinate system in the joint with an individual coordinate $\theta_i$, but these coordinates are unified as $(x, y)$ in the global coordinate system. 

The transformation function for each element is defined as
\begin{equation}
	\mathbf{x}_i = T_{\mathbf{q}}(\mathbf{q}_i) + \mathbf{x}_{Oi},
	\label{equ:tran1}
\end{equation}
where $\mathbf{x}_i$ and $\mathbf{q}_i$ are the position of $i$-th element in the global and local coordinate system, respectively. $\mathbf{x}_{Oi}$ is the global coordinate of $i$-th local coordinate system\textquotesingle s origin. 

If we take the derivative of equation (\ref{equ:tran1}) related to time, we will get
\begin{equation}
	\mathbf{\dot{x}}_i = \frac{\text{d}T_{\mathbf{q}}(\mathbf{q}_i)}{\text{d}t} \mathbf{\dot{q}}_i + \mathbf{\dot{x}}_{Oi} \coloneqq T_{\mathbf{\dot{q}}}(\mathbf{q}_i) \mathbf{\dot{q}}_i + \mathbf{\dot{x}}_{Oi}.
	\label{equ:tran2}
\end{equation}
We will utilise two multilayer perceptrons (MLPs) to separately model the transformation functions $T_{\mathbf{q}}$ and $T_{\mathbf{\dot{q}}}$ because of the completeness of MLPs \cite{hornik1989multilayer}. An alternative way is using one MLP to model $T_{\mathbf{q}}$ and differentiating the network to model $T_{\mathbf{\dot{q}}}$. If inputs are already in a global coordinate system, these MLPs will be two identity layers with $\mathbf{x}_0$ and $\mathbf{\dot{x}}_0$ being $\mathbf{0}$. 

\begin{figure}[htbp]
\vskip -0.07in
\centering
\includegraphics[width=1\linewidth]{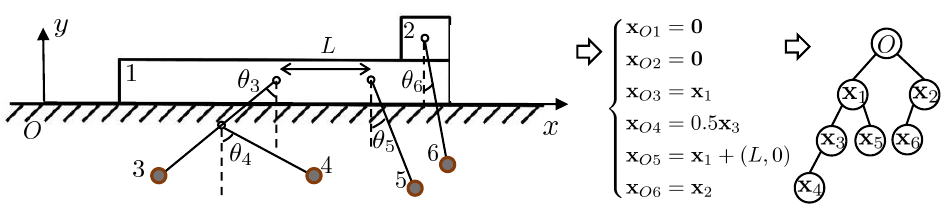}
\vskip -0.1in
\caption{An under-constrained system and the corresponding computation tree representing the calculation sequence. Here the origins of these elements are derived from root by traversal.}
\label{fig:tree}
\vskip -0.05in
\end{figure}

Though $(\mathbf{{x}}_{Oi}, \mathbf{\dot{x}}_{Oi})$ is unknown, we can construct a computation tree to compute the state of elements and origins from the root by traversal. As we assume, bijective connections between origins and elements can be obtained when local coordinate systems are constructed. Figure \ref{fig:tree} shows a calculation tree diagram of an example system.

This prior can be viewed as a loose constraint similar to \citealp{finzi2020simplifying}, where they directly introduced the holonomic constraints to simplify learning. However, imposing holonomic constraints is a stronger requirement, which involves detailed physical data. These constraints also treat dynamical functions as a whole, so the user cannot decouple elements from each other. 

There are two advantages of coordinate transformations. For potential energy, we can ensure a deep learning model works for all elements of the same type, as their inputs\textquotesingle\ forms are uniform. For kinetic energy, we can utilise expression $T_i = \sum_i \frac{1}{2} m_i \mathbf{\dot{x}}_i^{\intercal}\mathbf{\dot{x}}_i$ for each element, which is tenable as a property only in the global coordinate system. This property usually is not true in local coordinate systems because they can be moving.

\textbf{Energy Construction.} Establishing models for each element involves the kinetic and potential energy models. 

As discussed in Section \ref{sec:pre}, for each type of elements, we use two MLPs to separately model the potential energy between a element with the environment: $U_i = \text{NN}_U(\mathbf{x}_{i})$, and the potential energy among elements: $U_{ij} = \text{NN}_U(\mathbf{x}_{i}, \mathbf{x}_{j})$. As $U_{ij}$ and $U_{ji}$ are same, the system\textquotesingle s potential energy is:
\begin{equation}
	U = c_1\sum_i m_i U_i(\mathbf{x}_{i}) + c_2\sum_{i,j, i\ne j} \frac{m_im_j}{2} U_{ij}(\mathbf{x}_{i}, \mathbf{x}_{j}),
	\label{equ:U}
\end{equation}
where $c_1, c_2$ are learnable weights. For kinetic energy, the system\textquotesingle s kinetic energy $T_{\textrm{sys}}$ is the combination of each element\textquotesingle s kinetic energy $T_{i}$:
\begin{equation}
	T = \sum_i \frac{1}{2} m_i \mathbf{\dot{x}}_i^{\text{T}}\mathbf{\dot{x}}_i.
	\label{equ:T}
\end{equation}
This expression also follows the design in DeLaN \cite{lutter2018deep}. Here the mass matrix is diagonal as elements are independent. Since the mass for each element is its unique feature, it can be learnt as a parameter during training. 
If we want to introduce a new element into the framework, its mass should be parameterised or trained before. Overall, the mass will not affect the expression of the energy but only change the amplitude. 

\SetArgSty{textnormal}
\setlength{\textfloatsep}{0.16in}
\begin{algorithm}[tbp]
\DontPrintSemicolon
\KwIn{position $\mathbf{q}=(\mathbf{q}_1, ..., \mathbf{q}_n)$, velocity $\mathbf{\dot{q}}=(\mathbf{\dot{q}}_1,...,\mathbf{\dot{q}}_n)$, number of elements $n$, and number of inputs $m$ for $U$.}
\SetKwInOut{KwInit}{Variable}
\KwInit{origins $\mathbf{x}_{O}$, $\mathbf{\dot{x}}_{O}$, global coordinates $\mathbf{x}$, $\mathbf{\dot{x}}$, energies $U, T, L$, weights $c_1, c_2$.}
\SetKwProg{KwFnInit}{Initialise:}{}{end}
\KwFnInit{}{
$Tree \gets $ \Call{BuildComputationTree}{};\\
$\mathbf{x}_{O}, \mathbf{\dot{x}}_{O} = (\mathbf{x}_{O1}, ..., \mathbf{x}_{On}), (\mathbf{\dot{x}}_{O1}, ..., \mathbf{\dot{x}}_{On}) \gets \mathbf{0}, \mathbf{0}$;\\
$\mathbf{x}, \mathbf{\dot{x}} = (\mathbf{x}_1, ..., \mathbf{x}_n), (\mathbf{\dot{x}}_1, ..., \mathbf{\dot{x}}_n) \gets \mathbf{0}, \mathbf{0}$;\\
$U, T, L \gets 0, 0, 0$;\\
}
\For{$i$ \textbf{in} \Call{Iterate}{$Tree$}}{
$\mathbf{x}_{Oi}, \mathbf{\dot{x}}_{Oi} \gets$  \Call{UpdateOrigin}{$\mathbf{x}, \mathbf{\dot{x}}$}; \\
$\mathbf{x}_{i} \gets \textbf{T}_{\mathbf{q}}(\mathbf{q}_{i})+\mathbf{x}_{Oi}$; \\
$\mathbf{\dot{x}}_{i} \gets \textbf{T}_{\mathbf{\dot{q}}}(\mathbf{q}_{i}) \cdot \mathbf{\dot{q}_{i}} +\mathbf{\dot{x}}_{Oi}$;\\
}
\For{$i=1$ \textbf{to} $n$}{
$U \gets U + c_1 \cdot m_i U_i(\mathbf{x}_i)$;\\
\For{$j=1$ \textbf{to} $n$, $i \ne j$}{
$U \gets U + c_2 \cdot \frac{1}{2} m_i m_j U_{ij}(\mathbf{x}_i, \mathbf{x}_j)$;\\
}
}
\For{$i=1$ \textbf{to} $n$}{
$T \gets T + \sum_i \frac{1}{2} m_i \mathbf{\dot{x}}_i^{\text{T}}\mathbf{\dot{x}}_i$;\\
}
$L \gets T - U$;\\
$\mathbf{\ddot{q}} \gets \left(\frac{\partial^2 {L}}{\partial \mathbf{\dot{q}}^2}\right)^{-1} \left(\frac{\partial {L}}{\partial \mathbf{q}} - \frac{\partial^2 {L} }{\partial \mathbf{q}\partial {\mathbf{\dot{q}}}} \mathbf{\dot{q}}\right)$;\\
\Return{$\mathbf{\ddot{q}}$}
\caption{ModLaNet Framework}
\label{alg:framework}
\end{algorithm}

\textbf{Dynamics Derivation.} After constructing models for elements, we can introduce the physical inductive bias to derive the target value. We construct the Lagrangian of the system to describe the dynamical system following Equation (\ref{equ:lag}). Finally, We derive the dynamics with $\mathbf{{\ddot{q}}}$ following Equation (\ref{equ:results}). The framework to learn dynamics is also shown in Algorithm \ref{alg:framework}.

Here we have successfully modelled elements to construct the dynamic system using modularity. With modularity, we can reuse networks for many elements of the same type. The strengths are twofold.
For training, this feature reduces the parameters within our model, so we can utilise smaller datasets to train our model and will face more minor difficulties in convergence. 
For extension, we can train each type of elements individually, and then organise these models of different elements to construct more complex dynamical systems. Model reuse will provide more freedom for users in simulation.

\textbf{Examples.} We present two examples, multi-pendulum systems and multi-body systems, in Figure \ref{fig:Examples} to show how to implement the framework. 

A multi-pendulum system is $n$ single pendulums that are linked one by one using massless rods like Figure \ref{fig:Examples}(a). The Lagrangian of this system in polar coordinates is:
\begin{equation*}
\begin{split}
	{L}(\boldsymbol{\theta}, \boldsymbol{\dot{\theta}}) = & \left(\sum_{i=1}^{n}\sum_{j=1}^{i-1} (\frac{1}{2}m_il_j^2\dot{\theta}_j^2 + m_igl_j\cos{\theta_j}) \right.\\
	& \left. + \sum_{i=1}^{n}\sum_{j, k=1, j\ne k}^{i-1}m_il_jl_k\dot{\theta}_j\dot{\theta}_k\cos{(\theta_j-\theta_k)} \right) \text{,}
\end{split}
\end{equation*}
where $m_i$, $l_i$, $\theta_i$ and $\dot{\theta}_i$ are mass, pendulum length, angle position and angular velocity related to the $i$-th pendulum. $g$ is the gravitational acceleration.

For an multi-body system, $n$ particles move influenced by forces of attraction shown in Figure \ref{fig:Examples}(b). The Lagrangian is 
\begin{equation*}
	\begin{split}
	{L}(\mathbf{x},\mathbf{y},\dot{\mathbf{x}}, \dot{\mathbf{y}}) = &\sum_{i=1}^{n} \frac{1}{2}m_i (x_i^2 + y_i^2) \\
	& -\sum_{i=1}^{n} \sum_{j=1}^{i} \frac{G m_i m_j}{\sqrt{(x_i-x_j)^2+(y_i-y_j)^2}}\text{,}
	\end{split}
\end{equation*}
where $m_i$, $(x_i, y_i)$, and $(\dot{x}_i, \dot{y}_i)$ are mass, positions and velocities related to the $i$-th particle, $G$ is a physical constant. In both equations\textquotesingle\ right sides, the first line represents kinetic energy and the second relates to potential energy.

The framework is constructed as follows:
\begin{enumerate}
	\item \textbf{Coordinate Transformation.} For multi-pendulum systems, two 3-layer MLPs will be settled to separately model $\mathbf{T_{q}}$ and $\mathbf{T_{\dot{q}}}$. When construct coordinate systems, the origin of first pendulum is known fixed and is the same as the origin of the global coordinate system: $\mathbf{x}_{O1} = \mathbf{0}, \mathbf{\dot{x}}_{O1} = \mathbf{0}$. The origin of $i$-th pendulum $\mathbf{x}_{Oi}$ is linked to $(i-1)$-th pendulum recurrently: $\mathbf{x}_{Oi} = \mathbf{x}_{i-1}, \mathbf{\dot{x}}_{Oi} = \mathbf{\dot{x}}_{i-1}, i = 2, ..., n$. For multi-body systems, because states are already in global coordinate system, coordinate transformations are reduced to $\mathbf{x} = \mathbf{q}, \mathbf{\dot{x}} = \mathbf{\dot{q}}$.
	\item \textbf{Energy Construction.} To model potential energy in Equation (\ref{equ:U}), we use another two 3-layer MLPs: $U_i = \text{NN}_{U}(\mathbf{x}_i)$ and $U_{ij} = \text{NN}_{U}(\texttt{concatenate}(\mathbf{x}_i, \mathbf{x}_j))$. The kinetic energy is calculated from Equation (\ref{equ:T}). 
	\item \textbf{Dynamics Derivation.} Finally the dynamical system is derived with Equation (\ref{equ:lag}) and (\ref{equ:results}).
\end{enumerate}

\section{Experiments}
\begin{figure}[htbp]
\vskip -0.05in
\centering
\includegraphics[width=1\linewidth]{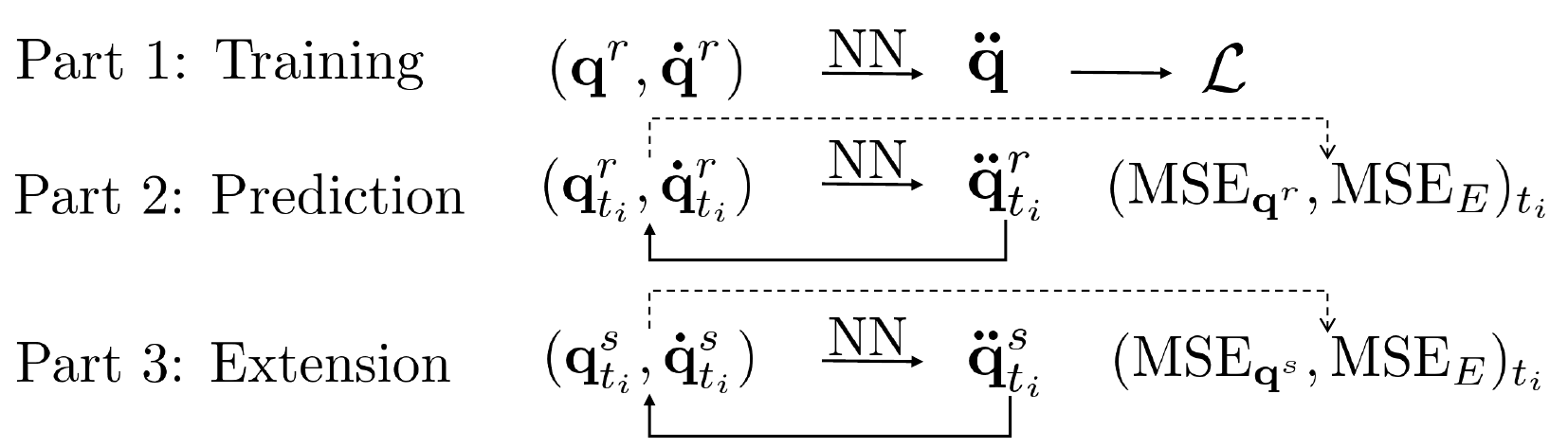}
\vskip -0.05in
\caption{How experiments work in each part. For part 1, we only do one-step forward propagation for the acceleration $\mathbf{\ddot{q}}$ to calculate loss $\mathcal{L}$ for training. For parts 2 and 3, we predict the trajectory for a range of time, which involves a multiple-step integration for estimating the accumulated MSEs related to positions and energies. $r$ and $s$ represent the dimensions of systems\textquotesingle\ states, and $s>r$ since dynamical systems in part 3 are more complex than in part 2.}
\label{fig:exp-steps}
\vskip -0.05in
\end{figure}

The target is to model the dynamics of multi-pendulum or multi-body systems discussed above. The whole experiment is divided into three parts shown in Figure \ref{fig:exp-steps}, the training, prediction and extension parts. Difficulties in simulation time and system dimensions gradually grow among these parts to demonstrate how each model behaves to learn the dynamics.

\fboxsep=3mm
\fboxrule=0.3pt

We include the two most representative methods for comparison, HNN and LNN \cite{Greydanus2019deep, cranmer2020lagrangian}. Most works mentioned in Section \ref{sec:works} are their applications or extensions, such as learning dynamics with control and processing image inputs (uncomparable in our experiments). The baseline model is a three-layer MLP directly modelling the evolution function $f$ in Equation (\ref{equ:system}) following HNN and LNN \cite{Greydanus2019deep, cranmer2020lagrangian}. Because parameters of our model are around $\sfrac{1}{8}$ of HNN and $\sfrac{1}{80}$ of LNN (shown in Table \ref{table:ExpResults}), we involve smaller versions called light HNN and light LNN for better comparison by reducing hidden dimensions.
 
%

\begin{table*}[htbp]
\setstretch{1.1}
\caption{Experiment settings and results for part 1 and 2}
\label{table:ExpResults}
\begin{sc}
\resizebox{1\textwidth}{!}{
\begin{tabular}{l|ccccc} 
\toprule
\multicolumn{4}{c}{Double-pendulum systems} &&\\
\midrule
Model & Params & Train Loss & Test Loss & Average Prediction $\text{MSE}_{\mathbf{q}}$ & Average Prediction $\text{MSE}_{E}$\\
\hline
ModLaNet & $6.2$k 
& $\mathbf{4.45\times 10^{-2} \pm 2.12\times 10^{-3}}$ 
& $\mathbf{3.47\times 10^{-2} \pm 5.70\times 10^{-3}}$ 
& $\mathbf{1.28\times 10^{0} \pm 1.25\times 10^{0}}$ 
& $\mathbf{4.60\times 10^{-2} \pm 5.02\times 10^{-2}}$  \\ \hline
Light LNN & $6.2$k 
& $2.60\times 10^{+3} \pm 2.79\times 10^{+2}$ 
& $4.26\times 10^{+3} \pm 4.72\times 10^{+3}$ 
& failed
& failed \\ \hline
Light HNN & $6.2$k 
& $3.97\times 10^{+2} \pm 2.80\times 10^{+1}$ 
& $5.00\times 10^{+2} \pm 7.35\times 10^{+1}$ 
& failed
& failed \\ \hline
LNN & 364.2k
& $9.11\times 10^{+2} \pm 3.97\times 10^{+1}$ 
& $7.11\times 10^{+2} \pm 6.99\times 10^{+1}$ 
& failed 
& failed \\ \hline
HNN & 43.2k
& $1.59\times 10^{+1} \pm 4.17\times 10^{0}$ 
& $9.77\times 10^{+3} \pm 1.03\times 10^{+2}$ 
& $4.91\times 10^{0} \pm 3.32\times 10^{0}$ 
& $1.24\times 10^{+1} \pm 2.37\times 10^{+1}$  \\ \hline
Baseline & 41.4k 
& $1.61\times 10^{-1} \pm 1.79\times 10^{-2}$
& $6.52\times 10^{+2} \pm 7.20\times 10^{+1}$
& $3.42\times 10^{0} \pm 1.92\times 10^{0}$
& $6.73\times 10^{+2} \pm 1.31\times 10^{+3}$ \\ 
\bottomrule
\toprule
\multicolumn{4}{c}{Three-body systems}  & &\\
\midrule
Model & Params & Train Loss & Test Loss & Average Prediction $\text{MSE}_{\mathbf{q}}$ & Average Prediction $\text{MSE}_{E}$\\
\hline
ModLaNet & $5.6$k 
& ${7.12\times 10^{-5} \pm 9.60\times 10^{-6}}$ 
& $\mathbf{1.56\times 10^{-5} \pm 8.59\times 10^{-7}}$  
& $\mathbf{3.96\times 10^{-2} \pm 2.94\times 10^{-2}}$  
& $\mathbf{1.10\times 10^{-3} \pm 1.61\times 10^{-3}}$   \\ \hline
Light LNN & $5.6$k 
& $1.12\times 10^{-3} \pm 6.01\times 10^{-5}$  
& $7.84\times 10^{-4} \pm 9.52\times 10^{-5}$ 
& failed
& failed \\ \hline
Light HNN & $5.6$k 
& $2.13\times 10^{-4} \pm 1.34\times 10^{-5}$ 
& $3.64\times 10^{-4} \pm 5.03\times 10^{-5}$
& $1.12\times 10^{-1} \pm 5.79\times 10^{-2}$ 
& $5.34\times 10^{-3} \pm 9.76\times 10^{-3}$  \\ \hline
LNN & 364.2k
& $\mathbf{4.84\times 10^{-6} \pm 7.15\times 10^{-7}}$ 
& $2.16\times 10^{-5} \pm 2.26\times 10^{-6}$ 
& failed 
& failed \\ \hline
HNN & 43.2k
& $2.09\times 10^{-4} \pm 1.28\times 10^{-5}$  
& $2.93\times 10^{-4} \pm 3.01\times 10^{-5}$ 
& $9.55\times 10^{-2} \pm 5.53\times 10^{-2}$ 
& $4.69\times 10^{-3} \pm 5.68\times 10^{-3}$  \\ \hline
Baseline & 43k 
& $2.56\times 10^{-4} \pm 2.04\times 10^{-5}$
& $1.45\times 10^{-4} \pm 1.17\times 10^{-5}$
& $2.64\times 10^{-1} \pm 1.12\times 10^{-1}$
& $2.08\times 10^{-2} \pm 2.25\times 10^{-2}$ \\ 
\bottomrule
\end{tabular}
}
\end{sc}
\end{table*}
\setstretch{1}

\subsection{Training} 
For training, instead of using single pendulum systems or two-body systems, we consider using the trajectories of chaotic double-pendulum systems or three-body systems. The reason is that these two systems are the two simplest chaotic systems among their groups \citep{shinbrot1992chaos, vaidyanathan2016advances}. The training will be more challenging to distinguish the best model from the counterparts but will not be so hard that no model can succeed.

During training, each model outputs the acceleration given input state $(\mathbf{q, \dot{q}})$. Then for optimisation we use $\text{L}_2$ loss:
\begin{equation}
	\mathcal{L} = \left\Vert \mathbf{\hat{\ddot{q}}} - \mathbf{\ddot{q}} \right\Vert_2,
	\label{equ:loss}
\end{equation}
where $\mathbf{\hat{\ddot{q}}}$ and $\mathbf{{\ddot{q}}}$ represent the acceleration obtained from the network and the ground truth, respectively. 

For double-pendulum systems, without loss of generality, the length and mass of each object are set to be 1 m and 1 kg. For three-body systems, the mass of each object is set to be 1 kg. Hyperparameters for tuning are learning rate ($10^{-4}$-$10^{-1}$), training epoch (1-20\textsc{k}), and activation functions. 

\textbf{Datasets.} The Lagrangian of target systems are hard coded related to the input state ($\mathbf{q}$, $\mathbf{\dot{q}}$). The corresponding derivatives are calculated using Autograd \citep{Autograd} to reconstruct $f$ in Equation (\ref{equ:system}). We then integrate it and obtain outputs $\mathbf{\ddot{q}}$ and inputs within a range of time $[t_0, t_g]$:
\begin{equation}
\begin{split}
	[(z_{t_1}, z_{t_2},..., z_{t_g})&,\,(\mathbf{\ddot{q}}_{t_0}, \mathbf{\ddot{q}}_{t_1},..., \mathbf{\ddot{q}}_{t_{g-1}})]= \\
	& \text{ODESolve}(z_{t_0}, f,(t_1, t_2, ..., t_g)),
	\label{equ:odesolve}
\end{split}
\end{equation}
where $z = (\mathbf{{q}}, \mathbf{\dot{q}})$ is the input state. SciPy provides the function \texttt{scipy.integrate.solve\_ivp} to numerically integrate ODEs \citep{2020SciPy-NMeth}. For HNN and baseline, their data follow Hamiltonian mechanics. The form is slightly different, but the calculation process is the same. 

For practical use in robotics and control, only small datasets are accessible in most cases, so we concern more about fast learning with small datasets in our work. $t_g$ is set to be 5 or 10s and the time step is 10 or 20 per second. The scale of the datasets is \textsc{10k-20k}. 

We summarise dataset sizes used in previous works. For HNN, the dataset size is close to ours, but they only consider single pendulum systems, where the dynamics are much simpler than double pendulum systems. For LNN, though they consider double pendulum systems, the dataset is enormous, corresponding to more parameters in LNN.

\begin{table}[htbp]
\centering
\setstretch{1}
\vskip -0.2in
\caption{Statistics of dataset sizes\protect\footnotemark[4].}
\vskip 0.05in
\label{table:datasetsize}
\begin{sc}
\resizebox{0.48\textwidth}{!}{
\begin{tabular}{c|cc|cc|c} 
\toprule
Model & \multicolumn{2}{c|}{ModLaNet} & \multicolumn{2}{c|}{HNN} & LNN \\
\hline
Task & 2-Pend & 3-Body & 1-Pend & 3-Body & 2-pend \\
\hline

\hline
Dataset Size &  10k & 20k & 2.25k & 10k & 307.2M\\
\bottomrule
\end{tabular}
}
\end{sc}
\vskip -0.1in
\end{table}
\setstretch{1}
\footnotetext[4]{Estimated from previous works \href{https://github.com/greydanus/hamiltonian-nn/tree/master/experiment-pend}{Hamiltonian Neural Networks} and \href{https://github.com/MilesCranmer/lagrangian_nns/blob/master/notebooks/DoublePendulum.ipynb}{Lagrangian Neural Networks}.}
 
\textbf{Training Results.} Training and testing results are shown in Table \ref{table:ExpResults}. Notice that test loss is slightly smaller than train loss. Because the tuning target $\mathcal{L}$  is optimised without normalisation, the amplitudes of $\mathbf{\ddot{q}}$ in datasets will influence the amplitude of losses. But this difference does not affect comparisons among methods based on the same dataset.

For double-pendulum systems, only our model indeed converges. The performance of baseline and HNN on the testing data shows that they overfit the training data but did not truly learn the evolution function. The loss of LNN is different from \citealp{cranmer2020lagrangian} where LNN achieves good performance ($10^{+2}$ versus $10^{-3}$). This difference can be explained by the use of the dataset, where the size of their dataset is significantly larger than ours (\textsc{300m} versus \textsc{10k}). Considering that LNN\textquotesingle s parameters are far larger than our model\textquotesingle s (\textsc{360k} versus \textsc{3.4k}), the training result means that LNN requires more data to converge to lower loss.

For three-body systems, all training processes converge, where our model achieves the lowest loss in testing (\sfrac{1}{10} of HNN and baseline, \sfrac{1}{2} of LNN). As in many cases where particles escape from forces of attraction with larger velocities, trajectories are not always chaotic. Therefore, it is easier to distil knowledge from the three-body systems than from double-pendulum systems.

The training on double-pendulum systems cannot converge for light HNN and light LNN. For three-body systems, ModLaNet outperforms them, implying overfitting is not the main cause for HNN/LNN\textquotesingle s inferior performance.
 
Overall, the training results imply that our framework can utilise lighter networks to capture the dynamical functions with limited data, which supports our advantages analysis in Section \ref{sec:method}.

\subsection{Prediction}
In the second part, we make a multiple-step prediction to examine the performance within a continuous period, which is different from training.
 Since errors will accumulate over time to affect simulation results, it is better to examine the stability and robustness of trained models.

The detailed procedure follows Equation (\ref{equ:odesolve}) within a period of time range $[t_0, t_h]$ ($t_h>t_g$) given an initial state $(\mathbf{q}_{t_0}, \mathbf{\dot{q}}_{t_0})$. In this part, $t_h$ is 30s. The evolution function $f$ is replaced trained models. After getting the simulation trajectory, the mean squared errors (MSEs) related to position and energy are to be calculated based on the trajectory of ground truth. This procedure will be repeated 100 times with arbitrary initialisations for the average and variance. 

We utilise the Runge-Kutta solver following the LNN and HNN\textquotesingle s setting. Choosing a solver maybe be necessary for models\textquotesingle\ stable performance in multi-step prediction. However, the main factor is how well the model is trained by one-step prediction without any ODE solver. 

\begin{figure}[tbp]
\vskip 0.05in
\centering
\includegraphics[width=1\columnwidth]{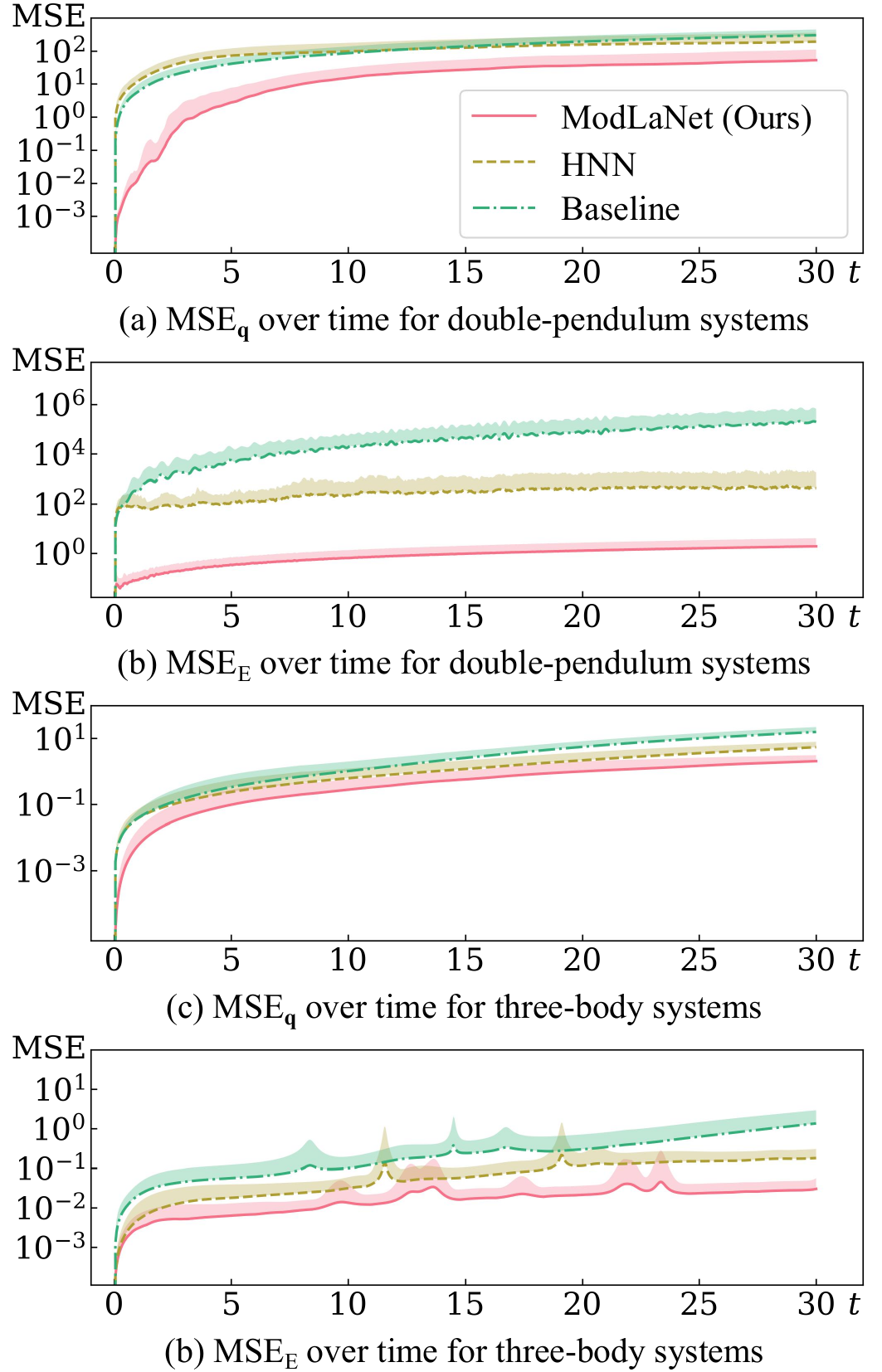}
\vskip -0.05in
\caption{The accumulated MSEs of position and energy over time using ModLaNet, HNN and baseline in the prediction part. (a-b) reveal the MSE of position and energy for double-pendulum systems, and (c-d) represent those for three-body systems. The shaded regions are the variance regions.}
\label{fig:exp-testing}
\vskip -0.05in
\end{figure}

\textbf{Prediction Results.}
We calculate MSEs of position and energy ($\text{MSE}_{\mathbf{q}}$ and $\text{MSE}_{E}$) related to time and each model. The results are shown in Figure \ref{fig:exp-testing} and Table \ref{table:ExpResults}. In Table \ref{table:ExpResults} we average these MSEs among time. Also, Figure \ref{fig:exp-trajectory} (a) and (b) visualise trajectories predicted by models.

For double-pendulum systems, the $\text{MSE}_{\mathbf{q}}$ of our model\textquotesingle s is $\sfrac{1}{4}$-$\sfrac{1}{3}$ of HNN and baseline. $\text{MSE}_{E}$ of our model is only $\sfrac{1}{250}$ of HNN, $\sfrac{1}{1500}$ of baseline. The main reason is that HNN and baseline can not converge to model the system with limited data. In contrast, our model can utilise light neural networks for regression after introducing the modularity, so our model performs better than others.

For three-body systems, our model\textquotesingle s $\text{MSE}_{\mathbf{q}}$ is $\sfrac{1}{3}$ of HNN and $\sfrac{1}{8}$ of baseline and $\text{MSE}_{E}$ is $\sfrac{2}{9}$ of HNN, $\sfrac{1}{20}$ of baseline. This result also reveals that our model achieved the most stable performance during a continuous-time and outperformed previous models.

Predictions using LNN failed in both tasks. The ODE solver could not correctly work because LNN outputted overlarge acceleration at specific regions. This outcome proves that LNN did not successfully learn the evolution function with limited data. 

\begin{figure}[tbp]
\vskip 0.05in
\centering
\includegraphics[width=1\columnwidth]{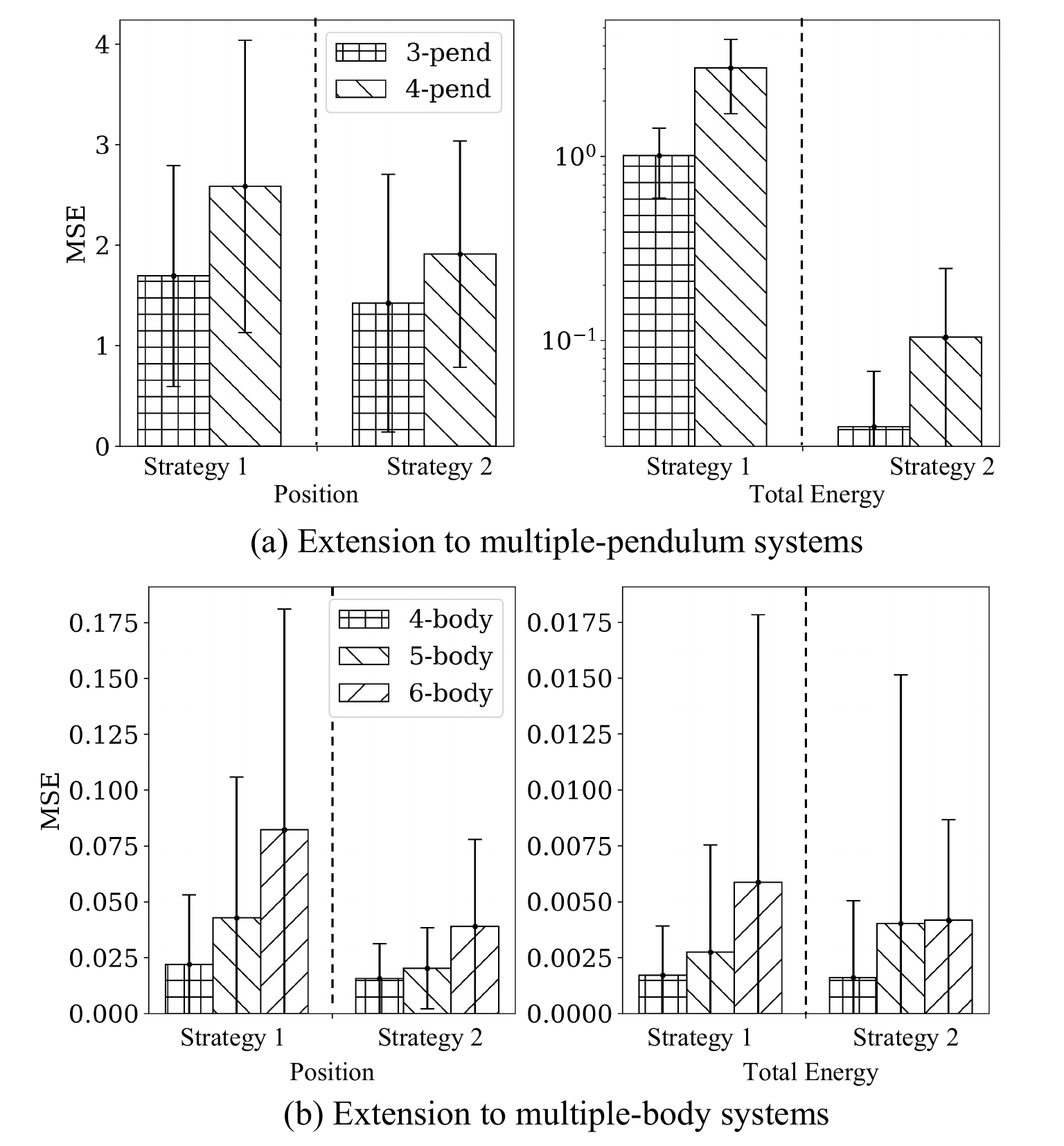}
\vskip -0.1in
\caption{Total MSEs of position and energy of ModLaNets using different strategies in the extension part. (a-b) list results of extensions to multi-pendulum and multi-body systems, respectively.}
\label{fig:exp-extension}
\vskip -0.05in
\end{figure}

\subsection{Extension} 
The third part examines the performance of our framework in extension. The procedure is the same as the last part but has complex cases: multiple-pendulum and multiple-object systems. The initial states are chosen randomly, and the prediction progress is repeated 100 times for average and variance. Prediction time $t_h$ is 10s. 

We evaluate our framework using two strategies: 1. We directly extend the trained framework to construct target systems and make predictions during a range of time. 2. Before extensions, we utilise minor datasets of multiple-pendulum or multiple-body systems to retrain the models for fine-tuning, similar to transfer learning. The size of the datasets is only $\sfrac{1}{10}$ of those used in the training part. And the retraining epoch is decreased to $\sfrac{1}{5}$.

\begin{figure}[H]
\vskip 0.1in
\centering
\includegraphics[width=0.95\columnwidth]{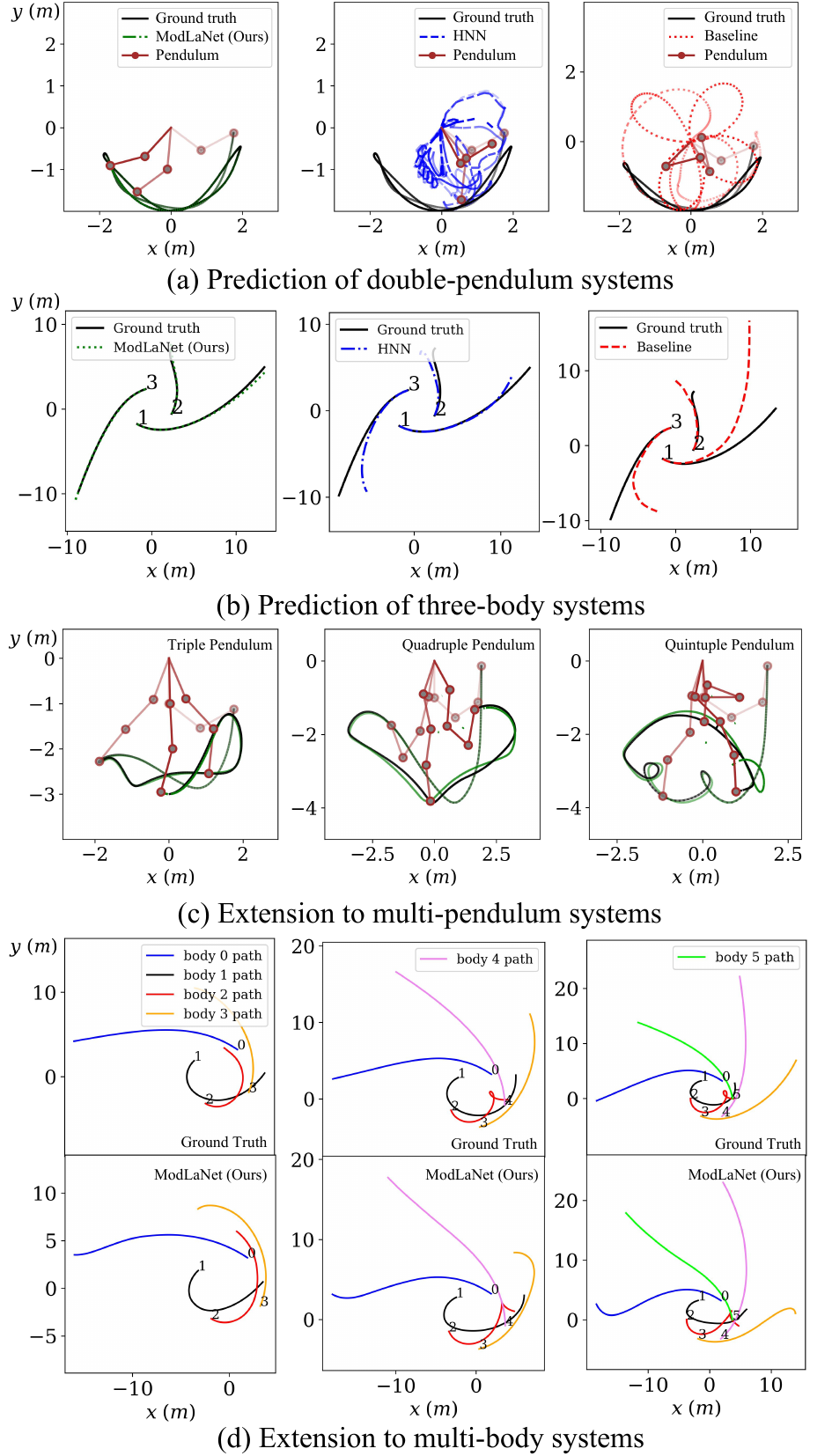}
\vskip -0.1in
\caption{Trajectories in the prediction and extension parts. (a-b) present trajectories of a double-pendulum system and a three-body system predicted by ModLaNet, HNN and baseline, respectively. (c-d) list how ModLaNet performed in multi-pendulum and multi-body systems in the extension part.}
\label{fig:exp-trajectory}
\vskip -0.1in
\end{figure}

\textbf{Extension Results.} The performance of our framework in extension to multi-body and multi-pendulum systems is shown in Figure \ref{fig:exp-extension}. Figure \ref{fig:exp-trajectory}(c)-(d) visualise trajectories as examples. Our models could predict reasonable trajectories obeying the laws of physics. 

For the first strategy, because we focus on modelling each element in the system, the MSEs grow related to the number of elements without retraining. 

For the second strategy, after retraining, the performance of our models becomes better. For multi-pendulum systems, $\text{MSE}_{\mathbf{q}}$ and $\text{MSE}_{E}$ are reduced by $16\%$-$26\%$ and $96\%$, respectively. For multi-body systems, the $\text{MSE}_{\mathbf{q}}$ decreases by $28\%$-$52\%$. The improvement in $\text{MSE}_{E}$ is not obvious as $\text{MSE}_{E}$ is already tiny. Results show that fine-tuning is a way to improve our models for more complex systems after training on simpler systems.

On the other hand, we observe cases where retraining can fail on over-complex systems with more data. For example, for 4th-pendulum systems, if we use the same amount of data as training, fine-tuning will fail with loss growing. Considering that it is more challenging than retraining if we directly train on multi-pendulum or multi-body systems, these cases imply that it can be too difficult to distil knowledge when dynamics functions are too complicated and chaotic. 

As a feasible way, instead of directly learning over-complex dynamical systems, modularity enables us to model related simpler systems with our framework, reducing training difficulties. After that, we can reorganise the trained network to construct the Lagrangian of over-complex dynamical systems with fine-tuning. Therefore, we believe that modularity and network reuse will significantly benefit deep learning models to conduct simulation tasks.

\section{Conclusion}
We propose a novel neural network framework called Modular Lagrangian Networks. This framework can model the generalisable dynamics using modularity and physical inductive bias. 
We focus on modelling elements in the system independently with modularity and capturing the laws of physics of the system with physical inductive bias. 
Our framework supports learning from simple systems with fewer elements and reusing the trained networks to build complicated systems with the same kind of elements.
Reusable networks can benefit in reducing difficulties in training, the scale of model and dataset. 
We estimate our framework by predicting the multi-pendulum and multi-body systems and evaluating models in three aspects, training, prediction and extension. Results show that our framework achieves the most stability in accuracy and data efficiency compared with our counterparts.
 
\textbf{Further Improvements.} 
1. We build two prototypes in multi-body and multi-pendulum systems as conceptual proposals. Each model contains only one type of element. We will extend our framework to support multiple types of elements together for modelling more systems, which is a natural expansion. 
2. We will update the models of $T$ and $U$ for general rigid-body systems, where the elements have geometry volume and cannot be viewed as mass points. 
3. Furthermore, all previous deep learning models mentioned in this paper cannot be utilised to model soft materials objects. The expansion to this field remains an open question. However, this question can be tackled in the view of modularity. Cooperated with the material point methods \cite{zhang2016material}, we will try to expand our frameworks to model soft material systems.

\section*{Acknowledgements}
This project is supported by HKSAR RGC GRF 11202119, 11207818, and HKSAR Technology Commission under the InnoHK initiative. Yupu Lu is sponsored by the HKU Presidential PhD Scholarship and the YS and Christabel Lung Postgraduate Scholarship. We would like to thank anounymous reviewers for their constructive suggestions.




\bibliography{ModLaNet}
\bibliographystyle{icml2022}

\newpage
\appendix
\onecolumn

\section{Experiment Supplementaries}
\label{Appendix:Arch}
%

\subsection{Part 1: Training}
Our model was built using PyTorch \citep{NEURIPS2019_9015}, and experiments were conducted in Ubuntu 20.04 using single core i7@3.7GHz. All models can run without the usage of GPU. Nonetheless, as the LNN model is too large compared with others, running it without GPU will be highly time-consuming. 

\begin{figure}[H]
\centering
\includegraphics[width=1\columnwidth]{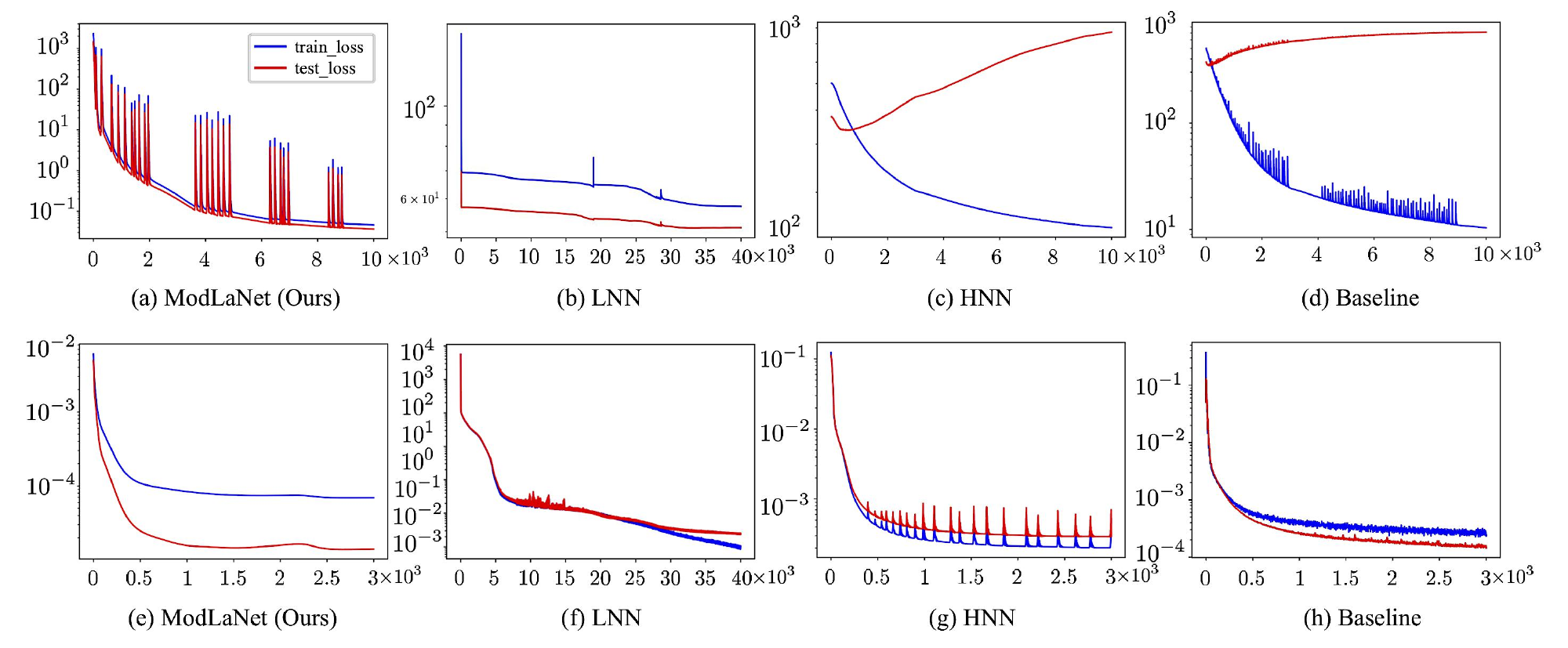}
\caption{The learning curves of four models in two experiments. (a-d) revealed the training results for double-pendulum systems, and (e-h) revealed those for three-body systems. }
\label{fig:exp-training}
\end{figure}

Figure \ref{fig:exp-training} shows the training results of four models in experiments. For double-pendulum systems, the training of HNN and baseline failed after long-time hyperparameter-tuning, where test losses continued increasing. Though LNN converged during training, the losses could not reach lower values. Only our model trained successfully. For three-body systems, all models converged to regress these systems. Our model achieved the lowest losses during training and testing. It should be pointed out that for LNNs, we followed the original $\text{L}_1$ loss in the training part, so the final training and testing losses are different from those ($\text{L}_2$ losses) listed in Table \ref{table:ExpResults}.

\subsubsection{Effects of Dataset Sizes}
We also investigate the effect of dataset sizes, which is the major difference in settings compared with counterparts\textquotesingle works. Only our model converges with small datasets (10-40\textsc{k}), which shows that our method outperforms others with better data efficiency. For large datasets, training is more complicated and more unstable because these datasets may influence the randomness of gradient descent for convergence and may require more training epochs and finer tuning. For practical use in robotics and control, only small datasets are accessible in most cases, so we concern more about fast learning with small datasets in our work.

\begin{table}[htbp]
\centering
\setstretch{1.2}
\vskip -0.15in
\caption{Different dataset sizes on double-pendulum systems.}
\vskip 0.05in
\label{table:data}
\begin{sc}
\resizebox{0.75\textwidth}{!}{
\begin{tabular}{l|ccccc|c|c} 
\toprule
Model & \multicolumn{5}{c|}{ModLaNet} & HNN & LNN \\
\hline
Dataset size & 10k & 20k & 40k & 100k & 1m &  10k-1m &  10k-1m \\
\hline

\hline
Train Loss 
& $4.45\times 10^{-2}$
& $3.29\times 10^{-2}$
& $8.61\times 10^{-2}$ 
& fail 
& fail 
& All fail
& All fail\\
\hline
Test Loss 
& $3.47\times 10^{-2}$ 
& $3.02\times 10^{-2}$
& $7.82\times 10^{-2}$ 
& fail 
& fail
& All fail
& All fail\\
\bottomrule
\end{tabular}
}
\end{sc}
\end{table}
\setstretch{1}

\subsubsection{Effects of Physical Inductive Bias}
As discussed in Section \ref{sec:method}, HNN/LNN with more physical inductive bias, e.g., $H = \mathbf{p}^{\intercal}\mathbf{p}/2 + U(\mathbf{q})$, is theoretically incorrect when inputs states are described within local coordinates systems, so the related ablation study is omitted.

\newpage
\subsection{Part 2: Prediction with Models}
Here we reveal how these trained models behaved during predictions. Since models are trained with a single step, it is unknown how well they perform at a continuous-time. We selected one case in the appendix, plotted their predicted trajectories, and compared them with the ground truth to better visualise their performance.

\begin{figure}[H]
\centering
\includegraphics[width=0.9\columnwidth]{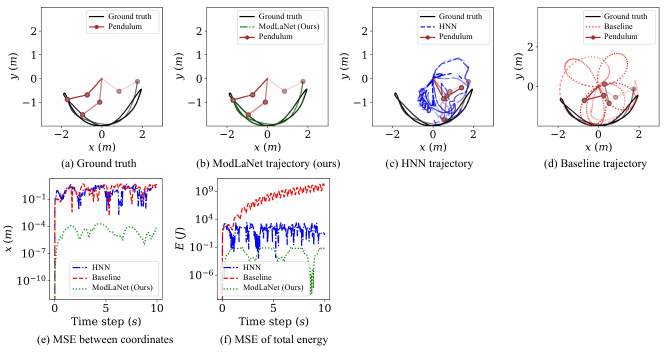}
\vspace*{-3mm}
\caption{Prediction results of a double pendulum system (best view in colour). Only our model (b) could make precise predictions close to the ground truth (a), while the HNN model (c) and baseline model (d) could not converge. MSEs related to our model in (e) and (f) weres much smaller than those of the other two models.}
\label{fig:exp-pend-2}
\end{figure}

\begin{figure}[htbp]
\centering
\includegraphics[width=0.9\columnwidth]{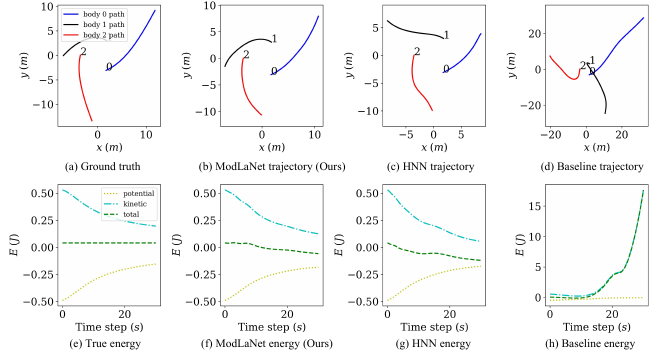}
\vspace*{-3mm}
\caption{Prediction results of a three-body system (best view in colour). 
Our prediction (b) here was closest to the ground truth (a). The trajectory of object one predicted by HNN (c) was away from the ground truth. The baseline performance was the worst, with the orbit (d) and energy (h) divergence. Our ModLaNet (f) and HNN model (g) were also able to predict the change of the energy, while our method could achieve more precise predictions of the ground truth (e).}
\label{fig:exp-body-3-short}
\end{figure}

\vspace*{-3mm}
\subsection{Part 3: Extension to Complex Systems}
\begin{figure}[H]
\centering
\includegraphics[width=1\columnwidth]{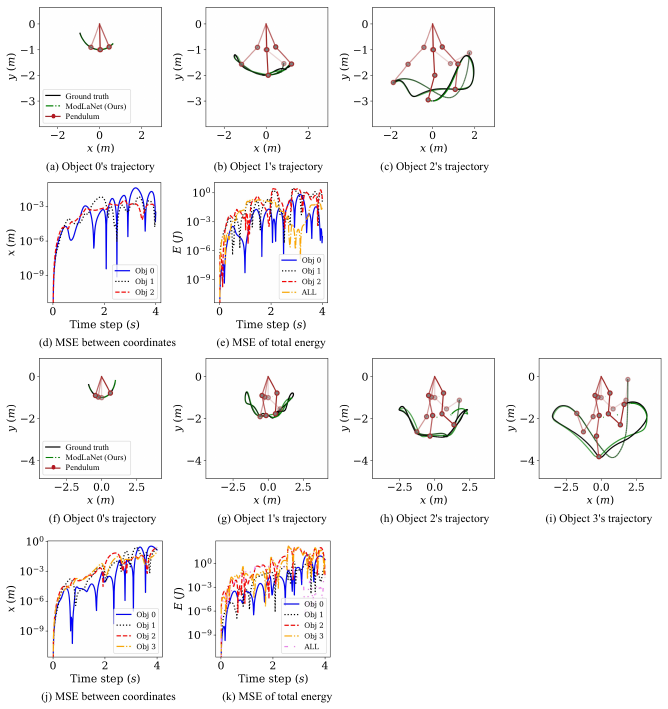}
\caption{Prediction results of triple-pendulum (a-e) and quadruple-pendulum (f-k) systems for extensions (best view in colour). The trajectories of the triple pendulum (a-c) and the quadruple pendulum (h-k) are listed compared with the ground truth. MSEs of different components shown in (d-g) indicate that predictions by our model were close to the ground truth.}
\label{fig:exp-pend-2+}
\end{figure}

\begin{figure}[htbp]
\centering
\includegraphics[width=1\columnwidth]{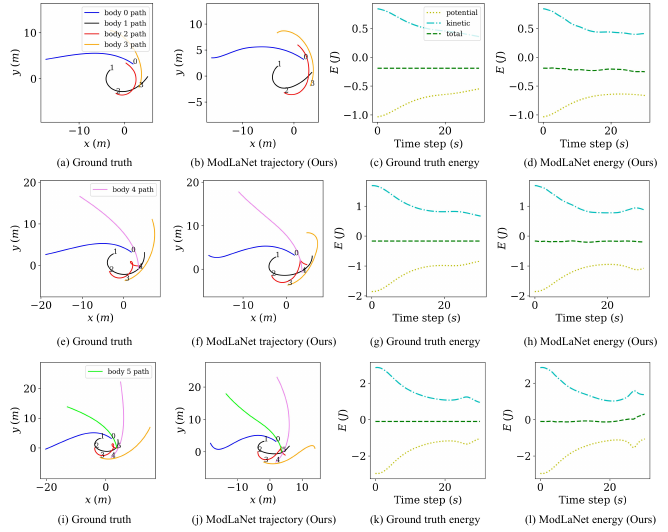}
\vspace*{-5mm}
\caption{Prediction results of multi-body ($>3$) systems for extrapolations (best view in colour). 
The trajectories corresponding to 4, 5, and 6-body systems (b, f, j) simulated using the ModLaNet model could grasp the main motion features similar to the ground truth (a, e, i). Our model could also predict energy change and retain the conservation of total energy (d, h, l) like the ground truth (c, g, k). As a result, our model could make great extensions to multi-body systems with the same law of physics, predict the correct motion and ensure the stability of the energy.}
\label{fig:exp-body-3+}
\end{figure}

Below are Figure \ref{fig:exp-pend-2+} and \ref{fig:exp-body-3+} showing how our ModLaNet performed at the extension to complex systems. We picked up cases of multi-pendulum systems and multi-body systems for visualisation.

\end{document}